\documentclass{article} 
\usepackage{nips15submit_e,times}
\usepackage{url}
\usepackage{latexsym}
\usepackage{float}
\usepackage{epsfig}
\usepackage{amsmath}
\usepackage{multirow}
\usepackage{graphicx}
\usepackage{amsthm}
\usepackage{amssymb}
\usepackage{natbib}
\usepackage{float}
\title{Reasoning about Linguistic Regularities in Word Embeddings using Matrix Manifolds}
\author{
Sridhar Mahadevan\\
University of Massachusetts Amherst\\
\texttt{mahadeva@cs.umass.edu} \\
\And
Sarath Chandar \\
IBM Research, USA \\
\texttt{apsarathchandar@gmail.com} \\
}

\nipsfinalcopy 

\begin{document}

\maketitle

\begin{abstract}
Recent work has explored methods for learning continuous vector space word representations reflecting  
 the underlying semantics of words. Simple vector space arithmetic using cosine distances has been shown to capture  certain types of analogies, such as reasoning about plurals from singulars, past tense from present tense, etc. In this paper, we introduce a new approach to capture analogies in continuous  word representations, based on modeling not just individual word vectors, but rather the {\em subspaces} spanned by groups of words. We exploit the property that the set of subspaces in $n$-dimensional Euclidean space form a curved manifold space called the {\em Grassmannian}, a quotient subgroup of the Lie group of rotations in $n$-dimensions. Based on this mathematical model, we develop a modified cosine distance model based on geodesic kernels that captures relation-specific distances across word categories. Our experiments on analogy tasks show that our approach performs significantly better than the previous approaches for the given task.
\end{abstract}

\section{Introduction}

In the past few decades, there has been growing interest in machine learning of continuous space representations of linguistic entities, such as words, sentences, paragraphs, and documents \cite{hinton:1986,elman:1990,bengio:2003,hinton:nips2008,mikolov:acl2013,mikolov:nips2013}. A recurrent neural network model was introduced in \cite{mikolov:2010}, and made widely available as the {\tt word2vec} program. It has been shown that continuous space representations learned by {\tt word2vec} were fairly accurate in capturing certain syntactic and semantic regularities, which could be revealed by relatively simple vector arithmetic \cite{mikolov:acl2013}. In one well-known example, Mikolov et al. \cite{mikolov:acl2013} showed that the vector representation of {\tt queen} could be inferred by a simple linear combination of the  vectors representing {\tt king}, {\tt man}, and {\tt woman} ({\tt king} - {\tt man} + {\tt woman}). However, the resulting vector might not correspond to vector representation of any of the words in the vocabulary. Cosine similarity   was used between the resultant vector and all word vectors to find the word in the voabulary that has maximum similarity with the resultant word. A more comprehensive study by  Levy and Goldberg \cite{goldberg-levy:2014} showed that a modified similarity metric based on multiplicative combination of cosine terms resulted in improved performance. 
A recent study by Levy et al. \cite{levy2015} verified the superiority of the modified similarity metric with several word representations. 

In this paper, we introduce a new approach to modeling word vector relationships. At the heart of our approach is the distinction that we model not just the individual words vectors, but rather the {\em subspaces} formed from groups of related words. For example, in inferring the plurals of words from their singulars, such as {\tt apples} from {\tt apple}, or {\tt women} from {\tt woman}, we model the subspaces of plural words as well as singular words. We exploit well-known mathematical properties of subspaces, including principally the property that the set of $k-$dimensional subspaces of $n$-dimensional Euclidean space forms a curved manifold called the {\em Grassmannian} \cite{edelman:siam-review}. It is well-known that the Grassmannian is a quotient subgroup of the {\em Lie} group of rotations in $n$-dimensions. We use these mathematical properties to derive a modified cosine distance, using which we obtain remarkably improved results in the same word analogy task studied previously \cite{,mikolov:acl2013,goldberg-levy:2014}.  

Recent work has developed efficient algorithms for doing inference on Grassmannian manifolds, and this area has been well explored in computer vision  \cite{gopalan:pami, gong:cvpr2012}. Gopalan et al.\cite{gopalan:pami} used the properties of Grassmannian manifolds to perform domain adaptation in Image classification by sampling subspaces between the source and target subspace on the geodesic flow between them. Geodesic flow is the shortest path between two points on curved manifolds. Gong et al. \cite{gong:cvpr2012} extended this idea by integrating over all subspaces in the geodesic flow from source to target subspace by computing the Geodesic Flow Kernel (GFK). 

In this paper, we  propose to develop a new approach to computing with word space embeddings by constructing a distance function based on constructing the geodesic flow kernel   between subspaces defined by various groups of words related by different relations, such as {\tt past-tense}, {\tt plural}, {\tt capital-of}, {\tt currency-of}, and so on. The intuitive idea is that by explicitly computing shortest-path geodesics between subspaces of word vectors, we can automatically determine a customized distance function on the Grassmannian manifold that specifically captures the way different relations map across word vectors, rather than assuming a simple vector translation model as in past work. As we will see later, the significant error reductions we achieve show that this intuition appears to be correct. 

The major contribution of this paper is the introduction of Grassmannian manifold based approach for reasoning in word embeddings. Even though this has been previously applied in image classification (a vision task), we demonstrate their success in learning analogies (an NLP task). This opens up several interesting questions for further research which we will describe at the end of the paper.



Here is a roadmap to the rest of the paper.  We begin in Section~\ref{word2vec} with a brief review of continuous space vector models of words. Section~\ref{analog-task} describes the analogical reasoning task. In Section~\ref{math}, we describe the proposed approach for learning relations using matrix manifolds. Section~\ref{experiments} describes the experimental results in detail, comparing our approach with previous methods. Section~\ref{future} concludes the paper by discussing directions for further research. 

\section{Vector Space Word Models}
\label{word2vec} 

Continuous vector-space word models have a long and distinguished history \cite{bengio:2003,elman:1990,hinton:1986,hinton:nips2008}. In recent years, with the popularity of so-called ``deep" learning" methods \cite{rbms}, the use of feedforward and recurrent neural networks in learning continuous vector-space word models has increased. The work of Mikolov et al. \cite{mikolov:acl2013,mikolov:nips2013,mikolov:2010} has done much to popularize this problem, and their {\tt word2vec} program has been used quite widely in a number of related studies \cite{goldberg-levy:2014,levy2015}. Recently, Levy et al., \cite{levy2015}, through a series of experiments, showed that traditional count based methods for word representation are not inferior to these neural based word representation algorithms. 

In this paper, we consider two word representation learning algorithms: Skip Grams with Negative Sampling (SGNS) \cite{mikolov:nips2013} and Positive Pointwise Mutual Information (PPMI) with SVD approximation. SGNS is a neural based algorithm while PPMI is a count based algorithm. In the Pointwise Mutual Information (PMI) based approach, words are represented by a sparse matrix M, where the rows corresponds to words in the vocabulary and the columns corresponds to the context. Each entry in the matrix corresponds to   PMI between the word and the context. We use Positive PMI (PPMI) where all the negative values in the matrix are replaced by 0. PPMI matrices are sparse and high dimensional. So we do truncated SVD to come up with dense vector representation of PPMI which is low dimensional. Levy and Goldberg \cite{levynips} showed that SGNS is implicitly factorizing a word context matrix whose cell's values are PMI, shifted by some global context.

\section{Analogical Reasoning Task} 
\label{analog-task} 

In the classic word analogy task studied in \cite{mikolov:acl2013,goldberg-levy:2014}, we are given two pairs of words that share a relation, such as {\tt man:woman} and {\tt king:queen}, or {\tt run:running} and {\tt walk:walking}. Typically, the identity of the fourth word is hidden, and we are required to infer it from the three given words. Assuming the problem is abstractly represented as $a$ is to $b$ as $x$ is to $y$, we are required to infer $y$ given the known identities of $a$, $b$, and $x$. 

Mikolov et al. \cite{mikolov:acl2013} proposed using a simple cosine similarity measure, whereby the missing word $y$ was filled in by solving the optimization problem
\begin{equation}
\label{mikolov}
\mbox{argmax}_{y \in V} \delta(\omega_y, \omega_x - \omega_a + \omega_b) 
\end{equation}
where $\omega_i$ is the vector space $D$-dimensonal embedding of word $i$ and $\delta$ is the cosine similarity given by 
\begin{equation}
\label{cos}
\delta(i,j) = \frac{\omega_i^T \omega_j}{\|\omega_i \|_2 \| \omega_j \|_2 } 
\end{equation} 

Let us call this method as CosADD. Levy and Goldberg  \cite{goldberg-levy:2014} proposed an alternative similarity measure using the same cosine similarity as Equation~\ref{cos}, but where the terms are used multiplicatively rather than additively as in Equation~\ref{mikolov}. Specifically, they proposed using the following multiplicative distance measure: 
\begin{equation}
\label{gl}
\mbox{argmax}_{y \in V} \frac{\delta(y,b) \delta(y, x)}{\delta(y, a) + \epsilon} 
\end{equation} 
where $\epsilon$ is some small constant (such as $\epsilon  = 0.001$ in our experiments). Let us call this method as CosMUL.

Our original motivation for this work stemmed from noticing that the simple vector arithmetic approach described in earlier work appeared to work well for some relations, but rather poorly for others. This suggested that the underlying space of vectors in the subspaces spanned by words that fill in $x$ vs. $y$ were rather non-homogeneous, and a simple universal rule such as vector subtraction or addition that did not take into account the specific relationship would do less well than one that exploited the knowledge of the specific relationship. Of course, such an approach is only pragmatic if the modified distance measure could somehow be automatically learned from training samples. In the next section, we propose one such approach.

\section{Reasoning on Grassmannian Manifolds}
\label{math}

\begin{figure}[ht]
\begin{center}
\begin{minipage}[t]{0.4\textwidth}
\includegraphics[width=\textwidth,height=2in]{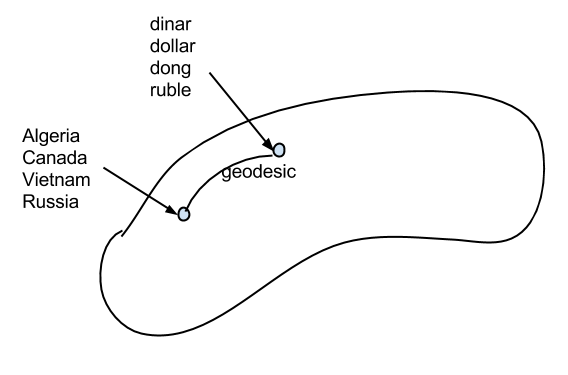}
\end{minipage}
\end{center} 
\vskip -0.2in
\caption{This figure illustrates the key idea underlying our subspace-based approach. A group of related word vectors are combined into a low-dimensional subspace, visualized by a small circle above, which represents a single point on the Grassmannian manifold. The  shortest-path geodesic distances between subspaces are explicitly  computed to generate a customized distance function for each relation. This figure illustrates the geodesic between countries and their currencies.}  
\label{geodesic} 
\end{figure}

Our approach builds on the key insight of explicitly representing the {\em subspaces} spanned by related groups of word vectors (see Figure~\ref{geodesic}). Given word vectors are embedded in an {\em ambient} Euclidean space of dimension $D$, we construct a low-dimensional representation of subspaces of size $d \ll D$, each representing groups of vectors. Given analogy tasks of the form {\tt  A is to B as X is to Y}, we construct subspaces from the list of sample training words comprising the categories defined by $A$ and $B$. For example, in the case of plurals, a sample word in the category $A$ is {\tt woman}, and a sample word in the category $B$ is {\tt women}. We use principal components analysis (PCA) to compute low-dimensional subspaces of size $d$, although any dimensionality reduction method could be used.  Many of the methods for constructing low-dimensional representations, from classic methods such as PCA \cite{jolliffe:pca} to modern methods such as Laplacian eigenmaps \cite{belkin:ml}, search for orthogonal subspaces of dimension $d \ll D$, the ambient dimension in which the dataset reside. A fundamental property of the set of all $d$-dimensional orthogonal subspaces in $\mathbb{R}^D$ is that they form a manifold, where each point is itself a subspace (see Figure~\ref{geodesic}). 


 Now, we need to compute the geodesic flow kernel which integrates over the geodesic flow between head subspace and tail subspace, so that we can project the word embeddings onto this relation specific kernel space. To compute the geodesic flow kernel, we need to compute the shortest path geodesic between two points on the Grassmannian manifold. In our setting, this corresponds to computing the shortest path geodesic between the points in the manifold which corresponds to the head subspace and the tail subspace.
 
Let the size of the word embeddings be $D$. Let $H$ denotes word embedding matrix where each row corresponds to word embedding of corresponding word in $A$ (head of analogy example) and $T$ denotes the word embedding matrix where each row corresponds to word embedding of corresponding word in $B$ (tail of analogy example). Now we learn $d$-dimensional subspaces for both $H$ and $T$. Let $P_H, P_T \in \mathbb{R}^{D \times d}$ denote the two sets of basis vectors that span the subspaces for the ``head" and `` tail" for a relation (for example, words and their plurals, or past and present tenses of verbs, and so on). Let $R_H \in \mathbb{R}^{D \times (D-d)}$ be the orthogonal complement to the subspace $P_H$, such that $P^T_H R_H = 0$. The geodesic flow shortest path between two points $P_H$ and $P_T$ of a Grassmannian Lie group can be parameterized by a one parameter exponential flow $\Phi(t)  = P_H exp(t B) P_T$ such that $\Phi(0) = P_H$, and $\Phi(1) = P_T$ and where $B$ is a skew-symmetric matrix and $exp$ refers to matrix exponential. For any other point $t$ other than $0$ or $1$, the flow $\Phi(t)$ can be computed as:
 \begin{equation} 
\Phi(t) = P_H U_1 \Gamma(t) - R_H U_2 \Sigma(t),
\end{equation} 
where $U_1 \in \mathbb{R}^{d \times d}$ and $U_2 \in \mathbb{R}^{(D - d) \times d}$ are orthonormal (length-preserving rotation) matrices that can be computed by a pair of singular value decompositions (SVD) as follows: 
\begin{equation}
\label{svd} 
P^T_H P_T = U_1 \Gamma V^T, \ \ \  R^T_H P_T = -U_2 \Sigma V^T 
\end{equation}
The $d \times d$ diagonal matrices $\Gamma$ and $\Sigma$ are particularly important since they represent $\cos(\theta_i)$ and $\sin(\theta_i), i = 1, \ldots, d$, where $\theta_i$  are the so-called {\em principal angles} between the subspaces $P_H$ and $P_T$.

\begin{figure}[H]
\begin{center}
\begin{minipage}[t]{0.4\textwidth}
\includegraphics[width=\textwidth,height=2in]{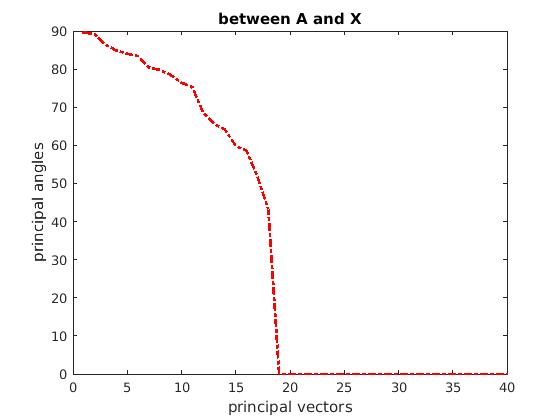}
\end{minipage}
\begin{minipage}[t]{0.4\textwidth}
\includegraphics[width=\textwidth,height=2in]{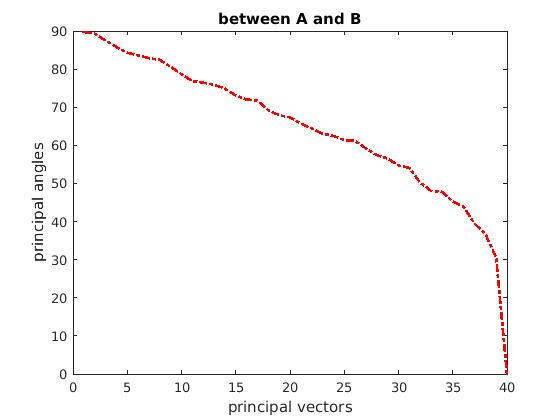}
\end{minipage}
\end{center} 
\caption{This figure illustrates the principal angles between two pairs of subspaces involved in family relationships. On left is a plot of principal angles (ranging from a maximum of $90$ degrees to a minimum of $0$ degrees) between subspaces $A$ and $X$, and on the right is plotted the principal angles between the subspaces $A$ and $B$ in a training set of analogical family relationships of the form $A$ is to $B$ as $X$ is to $Y$. The horizontal axis measures the dimension of the induced low-dimensional subspace.} 
\label{family-pa} 
\end{figure}

Figure~\ref{family-pa} illustrates a pair of subspaces involved in family relationships, and the principal angles between them. Note that the maximum angle between two subspaces is $90$ degrees, and the subspaces get closer as the principal angles get closer to $0$. What this intuitively means is that the principal angles represent the degree of overlap between the subspaces, so that as the corresponding principal vectors are added to each subspace the degree of overlap between the two subspaces increases. As Figure~\ref{family-pa} shows, the degree of overlap between the subspaces $A$ and $X$  increases much more quickly (causing the largest principal angle to shrink to $0$) than that between $A$ and $B$, as we would expect, because both $A$ and $X$ represent the ``head" in a family relationship.

Now let us describe how to compute the geodesic flow kernel $G_R$ specific to relation $R$. The basic idea is as follows. Each subspace $\Phi(t)$ along the curved path from the head $P_H$ to the tail $P_T$ represents a possible concept that lies ``in between" the subspace $A$ and $B$ (for example, $A$ and $B$ could represent ``singular" and ``plural" forms of a noun ). To obtain the projection of a word vector $x_i$ on a subspace $\Phi(t)$, we can just compute the dot product $\Phi(t)^T x_i$. Given two $D$-dimensional word vectors $x_i$ and $x_j$, we can simultaneously compute their projections on all the subspaces that lie between the ``head" and ``tail" subspaces by forming the geodesic flow kernel \cite{gong:cvpr2012}, defined as
\begin{equation}
\langle z_i, z_j \rangle_R = \int_0^1 (\Phi(t)^T_R x_i)^T (\Phi(t)^T_R x_j) dt = x^T_i G_R x_j 
\end{equation} 
 
 The geodesic kernel matrix $G_R$ can be computed in closed form from the above matrices computed previously in Equation~\ref{svd} using singular value decomposition: 
 \begin{equation}
 G_R = \left( \begin{array}{cc} P_S U_1 & R_S U_2 \end{array} \right) \left( \begin{array}{cc} \Lambda_1 & \Lambda_2 \\ \Lambda_2 & \Lambda_3 \end{array} \right) \left( \begin{array}{c} U^T_1 P^T_S  \\ U^T_2 R^T_S \end{array} \right) 
 \end{equation} 
 where $\Lambda_i$ are diagonal matrices whose elements are given by: 
 \begin{equation}
 \lambda_{1i} (\lambda_{3i}) = 1 + (-)  \frac{\sin(2 \theta_i)}{2 \theta_i}, \lambda_{2i} = \frac{\cos(2 \theta_i) - 1}{2 \theta_i}  
 \end{equation}
 A more detailed discussion of geodesic flow kernels can be found in \cite{gopalan:pami,gong:cvpr2012}, which applies them to problems in computer vision. This is the first application of these ideas to natural language processing, to the best of our knowledge. 

Once we have the relation specific GFKs computed, now we can perform our analogy task in the kernel space. The modified cosine distance would be,

\begin{equation} 
\label{gfk}
\delta_{G_R}(i, k) = \frac{\omega_i^T G_R \omega_k}{\| \sqrt{G_R} \omega_i \|_2 \| \sqrt{G_R} \omega_k \|_2}
\end{equation} 
Here, $\delta_{G_R}$ defines the modified cosine distance between word vectors $\omega_i$ and $\omega_k$ corresponding to words $i$ and $k$ for relation $R$ using a {\tt kernel} $G$, which captures the specific way in which the standard distance between categories must be modified for relation $R$.  Unlike the standard cosine distance, which treats each dimension equivalently, our approach automatically learns to weight the different dimensions adaptively from training data to customize it to different relations. The kernel $G$ is a positive definite matrix, which is learned from samples of word relationships.  

Now, similar to CosADD, we can define GFKCosADD,
 
\begin{equation}
\label{newmikolov}
\mbox{argmax}_{y \in V} \delta_{G_R}(\omega_y, \omega_x - \omega_a + \omega_b) 
\end{equation}
where $\omega_i$ is the vector space $D$-dimensonal embedding of word $i$ and $\delta_{G_R}$ is the modified cosine similarity given by \ref{gfk}. We can also compute GFKCosMUL (CosMUL in the kernel space) as: 

\begin{equation}
\label{gl}
\mbox{argmax}_{y \in V} \frac{\delta_{G_R}(y,b) \delta_{G_R}(y, x)}{\delta_{G_R}(y, a) + \epsilon} 
\end{equation} 
where $\epsilon$ is some small constant.

\section{Experiments} 
\label{experiments} 

In this section, we will describe the experimental results on Google and MSR analogy datasets. We learn word embeddings using two different learning algorithms : SGNS and SVD approximation of PPMI. We perform the analogy task using four distance metrics: two relation-independent metrics, CosADD and  CosMUL, and two relation-specific metrics, GFKCosADD and  GFKCosMUL. Our primary goal is to investigate the potential reduction in error rate when we learn relation specific kernels, as compared to using relation-independent metrics, CosADD and CosMUL. 

\subsection{Dataset} 

All word representation learning algorithms were trained on English Wikipedia (August 2013 dump), following the preprocessing steps mentioned in \cite{levy2015}. Words that appeared less than 100 times in the corpus were ignored. After preprocessing, we ended up with vocabulary of 189,533 terms. For SGNS we learn 500 dimensional representations. PPMI learns a sparse high dimensional representation which is projected to 500 dimensions using truncated SVD.

For the analogy task, we used the Google  and MSR datasets. The MSR dataset contains 8000 analogy questions. They are broadly classifed as : adjective, noun, and verb based questions. The Google dataset contains 19544 questions. It contains 14 relations. Out of vocabulary words were removed from both datasets.

\subsection{Experimental Setting}

For all the three word representation algorithms, we consider two important hyperparameters that might affect the quality of the representations learnt: window size of the context, and  positional context. We try both narrow and broad windows (2 and 5). When positional context is True, we consider the position of the context words as well, while we ignore the position when this parameter is set to False. This results in four possible settings. All the other hyperparameters of these two algorithms where set to default values as suggested by Levy et al. \cite{levy2015}.

We report accuracy in Google and MSR datasets in Table~\ref{t1} and Table~\ref{t2}, respectively. The results are micro-averaged over all relations in the dataset.

\begin{table}[h]
\begin{center}
\begin{tabular}{l|l|l|l|l|l}
\hline
Config &  Model &  CosADD & CosMUL & GFKCosADD & GFKCosMUL \\ \hline
\multirow{2}{*}{\begin{tabular}[c]{@{}l@{}}$win$=2,\\ $pos$=True\end{tabular}}  & SGNS        &  45.15\%            &    54.27\%          &          57.62\%       &        \textbf{62.35\%}         \\ 
                                                                           & SVD         &  43.66\%            &    60.05\%          &      58.66\%           &         \textbf{65.91\%}        \\ \hline
\multirow{2}{*}{\begin{tabular}[c]{@{}l@{}}$win$=5,\\ $pos$=True\end{tabular}}  & SGNS        &  53.17\%            &    62.19\%          &      67.68\%           &     \textbf{71.70\%}            \\ 
                                                                           & SVD         &  52.14\%            &    71.34\%          &   62.46\%               &    \textbf{74.18\%}             \\ \hline
\multirow{2}{*}{\begin{tabular}[c]{@{}l@{}}$win$=2,\\ $pos$=False\end{tabular}} & SGNS        &  49.41\%            &    63.21\%          &      71.17\%           &     \textbf{76.01\%}            \\ 
                                                                           & SVD         &  50.87\%            &    65.82\%          &      67.11\%           &  \textbf{72.45\%}                \\ \hline
\multirow{2}{*}{\begin{tabular}[c]{@{}l@{}}$win$=5,\\ $pos$=False\end{tabular}} & SGNS        &  56.14\%            &    74.43\%          &      81.06\%           &     \textbf{84.64\%}            \\  
                                                                           & SVD         &  60.82\%            &    75.14\%          &      72.29\%           &        \textbf{79.15\%}         \\ \hline
\end{tabular}
\caption{Accuracy obtained by various similarity measures in Google dataset. $win$ refers to the window size. $pos$ is True if position of the context is considered and False otherwise.}
\label{t1}
\end{center}
\end{table}

\begin{table}[h]
\begin{center}
\begin{tabular}{l|l|l|l|l|l}
\hline
Config &  Model &  CosADD & CosMUL & GFKCosADD & GFKCosMUL \\ \hline
\multirow{2}{*}{\begin{tabular}[c]{@{}l@{}}$win$=2,\\ $pos$=True\end{tabular}}  & SGNS        &      59.55\%        &    66.49\%          &      66.76\%           &        \textbf{68.36\%}         \\ 
                                                                           & SVD         &  50.59\%             &       65.38\%       &     59.11\%            &    \textbf{69.00\%}             \\ \hline
\multirow{2}{*}{\begin{tabular}[c]{@{}l@{}}$win$=5,\\ $pos$=True\end{tabular}}  & SGNS        &  61.39\%            &    69.66\%          &      71.42\%           &            \textbf{73.25\%}     \\ 
                                                                                                                                      & SVD         &  53.59\%            &    70.59\%          &      60.84\%           &    \textbf{72.18\%}             \\ \hline
\multirow{2}{*}{\begin{tabular}[c]{@{}l@{}}$win$=2,\\ $pos$=False\end{tabular}} & SGNS        &      59.41\%        &    69.87\%          &      72.70\%           &     \textbf{74.52\%}            \\ 
                                                                                                                                      & SVD         &  51.68\%            &    64.47\%          &      61.99\%           &         \textbf{66.25\%}        \\ \hline
\multirow{2}{*}{\begin{tabular}[c]{@{}l@{}}$win$=5,\\ $pos$=False\end{tabular}} & SGNS        &  64.48\%            &    76.00\%          &      78.81\%           &    \textbf{78.95\%}             \\  
                                                                                                                                      & SVD         &   52.50\%           &    \textbf{69.92\%}          &  62.25\%               &    67.05\%             \\ \hline
\end{tabular}
\caption{Accuracy obtained by various similarity measures in MSR dataset. $win$ refers to the window size. $pos$ is True if position of the context is considered and False otherwise.}
\label{t2}
\end{center}
\end{table}

From the tables, it is clear that GFK based similarity measures perform much better than respective non-GFK based similarity measures in most of the cases. We also report the relation-size accuracy in both the datasets in Table 3. Except for \texttt{captial-world} relation (where CosMUL performs better), GFK based approaches perform significantly better than Euclidean cosine similarity based methods.

\begin{table}[h]
\begin{tabular}{l|l|l|l|l|l}
\hline
                     & Relation                    & CosADD & CosMUL & GFKCosADD & GFKCosMUL \\ \hline
\multirow{14}{*}{Google} & capital-common-countries  &  89.52\% & 98.22\% & \textbf{100\%}     &  \textbf{100\%}  \\  
                     & capital-world               &  51.25\% & \textbf{80.43\%}  & 72.61\%  &  76.68\%    \\  
                     & city-in-state               & 7.62\% & 43.12\%  & 46.00\%      &  \textbf{69.59\%}  \\  
                     & currency                    & 18.57\% & 15.17\%     & \textbf{33.43\%}     & 27.86\% \\  
                     & family (gender inflections) & 69.36\% & 81.42\%    & \textbf{94.26\%}      & 93.67\%  \\  
                     & gram1-adjective-to-adverb   & 30.54\%& 39.91\%     &  \textbf{89.31\%}  &    86.18\%   \\  
                     & gram2-opposite              & 39.40\%& 45.32\%      & \textbf{75.00\%} &    73.02\%       \\  
                     & gram3-comparative           & 73.49\%& 88.81\%  & \textbf{92.71\%}       & 91.96\%       \\  
                     & gram4-superlative           & 33.80\% & 67.61\%     & 86.17\%       &    \textbf{90.43\%}     \\  
                     & gram5-present-participle    & 80.01\%& 92.32\%      & \textbf{99.81\%}       & 99.71\%       \\  
                     & gram6-nationality-adjective & 92.49\%& 95.30\%  & \textbf{98.93\%}       & 98.43\%      \\  
                     & gram7-past-tense            & 84.29\%& 93.79\%      & \textbf{99.80\%}       & 99.29\%       \\  
                     & gram8-plural (nouns)        & 80.03\%& 90.16\%      &    \textbf{98.19\%}   & 97.67\%       \\  
                     & gram9-pluran-verbs          & 82.52\%& 91.72\%      & \textbf{97.81\%}       & 97.58         \\ \hline
\multirow{3}{*}{MSR} & adjectives                  & 35.90\%& 47.19\%      & 59.55\%       & \textbf{60.44\%}       \\  
                     & nouns                       & 69.91\%& 83.04\%      & \textbf{84.10\%}      & 83.90\%          \\  
                     & verbs                       & 81.26\%& 91.86\%      & \textbf{89.03\%}       & 88.86\%           \\ \hline
\end{tabular}
\caption{Relation wise accuracy in Google and MSR datasets. Representations are learnt using SGNS with $win$=5 and $pos$=False. GFK-based methods perform better than their non-GFK based counterparts in all but one relation type.}
\label{t3}
\end{table}

Table 4 and Table 5 reports average rank of the of the correct answer in the ordered list of predictions made by the models. Ideally, this should be 1. These tables again demonstrate the superiority of GFK based approaches. We can see average rank for GFK based methods are significantly lower that their non-GFK based counterparts in most of the cases.

\begin{table}[H]
\begin{center}
\begin{tabular}{l|l|l|l|l|l}
\hline
Config &  Model &  CosADD & CosMUL & GFKCosADD & GFKCosMUL \\ \hline
\multirow{2}{*}{\begin{tabular}[c]{@{}l@{}}$win$=2,\\ $pos$=True\end{tabular}}  & SGNS        &     262.81         &    178.46          &   214.28              &     \textbf{149.42}            \\ 
                                                                                                                                      & SVD         &  332.73            &  128.01            &     279.41            &     \textbf{108.53}            \\ \hline
\multirow{2}{*}{\begin{tabular}[c]{@{}l@{}}$win$=5,\\ $pos$=True\end{tabular}}  & SGNS        &     165.69         &    116.81          &  124.67               &  \textbf{86.46}               \\ 
                                                                                                                                      & SVD         &  255.38            &      74.71        &      225.87           &    \textbf{64.35}             \\ \hline
\multirow{2}{*}{\begin{tabular}[c]{@{}l@{}}$win$=2,\\ $pos$=False\end{tabular}} & SGNS        &     110.74         &    74.94          &    83.36             &  \textbf{53.19}                \\ 
                                                                                                                                      & SVD         &   196.47           &  98.14            &      149.58           &      \textbf{76.38}           \\ \hline
\multirow{2}{*}{\begin{tabular}[c]{@{}l@{}}$win$=5,\\ $pos$=False\end{tabular}} & SGNS        &     60.03         &     41.61         &     39.25            &   \textbf{28.05}              \\  
                                                                                                                                      & SVD         &   116.65           &          61.53    &      101.03           &        \textbf{53.00}         \\ \hline
\end{tabular}
\caption{Average Rank obtained by various similarity measures in Google dataset. $win$ refers to the window size. $pos$ is True if position of the context is considered and False otherwise.}
\end{center}
\label{t4}
\end{table}

\begin{table}[H]
\begin{center}
\begin{tabular}{l|l|l|l|l|l}
\hline
Config &  Model &  CosADD & CosMUL & GFKCosADD & GFKCosMUL \\ \hline
\multirow{2}{*}{\begin{tabular}[c]{@{}l@{}}$win$=2,\\ $pos$=True\end{tabular}}  & SGNS        &         18.14     &         13.41     &     16.10            &   \textbf{12.33}              \\ 
                                                                                                                                      & SVD         &  23.51            &   15.38           &   21.84              &        \textbf{12.45}         \\ \hline
\multirow{2}{*}{\begin{tabular}[c]{@{}l@{}}$win$=5,\\ $pos$=True\end{tabular}}  & SGNS        &     13.68         &         11.26     &     12.37            &  \textbf{10.60}               \\ 
                                                                                                                                      & SVD         &       20.90       &   11.34           &       22.03          &    \textbf{ 11.33}            \\ \hline
\multirow{2}{*}{\begin{tabular}[c]{@{}l@{}}$win$=2,\\ $pos$=False\end{tabular}} & SGNS        &     11.73         &     8.89         &      10.45           &   \textbf{8.32}              \\ 
                                                                                                                                      & SVD         &  19.07            &   \textbf{14.29}           &       19.55          &        14.38         \\ \hline
\multirow{2}{*}{\begin{tabular}[c]{@{}l@{}}$win$=5\\ $pos$=False\end{tabular}} & SGNS        &      8.17        &   6.77           &        8.06         &  \textbf{7.31}               \\  
                                                                                                                                      & SVD         &   14.85           &  \textbf{9.14}            &        15.88         &      11.13           \\ \hline
\end{tabular}
\caption{Average Rank obtained by various similarity measures in MSR dataset. $win$ refers to the window size. $pos$ is True if position of the context is considered and False otherwise.}
\end{center}
\label{t5}
\end{table}

\begin{figure}[H]
\begin{center}
\begin{minipage}[t]{0.4\textwidth}
\includegraphics[width=\textwidth,height=2in]{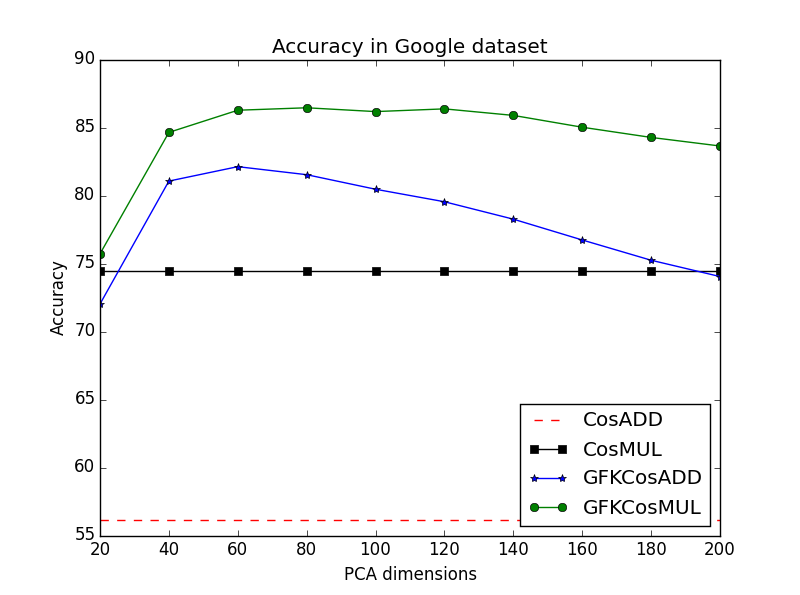}
\end{minipage}
\begin{minipage}[t]{0.4\textwidth}
\includegraphics[width=\textwidth,height=2in]{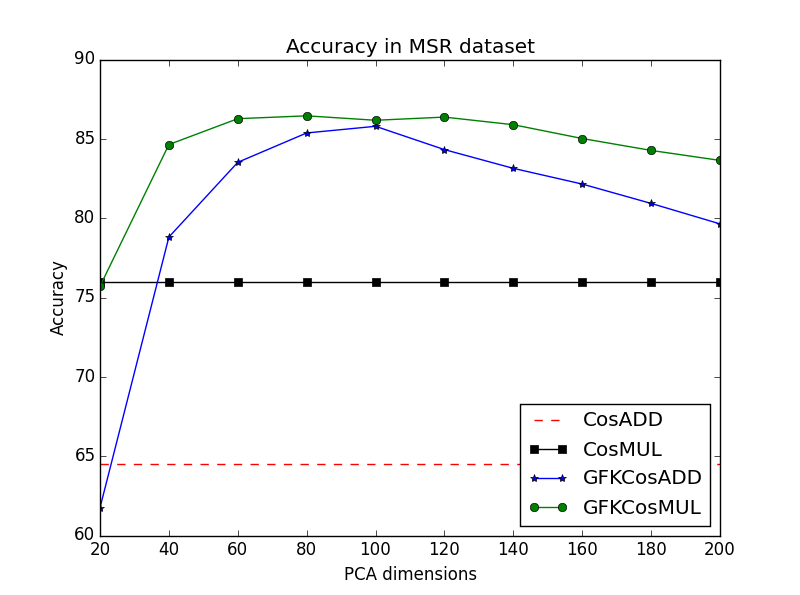}
\end{minipage}
\end{center} 
\caption{This figure explores the performance of the proposed GFK based methods on the Google dataset (left) and MSR dataset (right), both with varying subspace dimension from $d=20$ to $d=200$ in (steps of 20) compared to the fixed performance of the non-GFK based methods.} 
\label{pca-dimn} 
\end{figure}

An interesting question is how the performance of the GFK based methods varies with the dimensionality of the subspace embedding.  All the results in the above tables for our proposed GFK method are based on reducing the dimensionality of word embedding from the original $D = 500$ to a subspace of dimension $d=40$. Figure~\ref{pca-dimn} plots the performance of the GFK based methods and the previous methods on the Google dataset and MSR dataset, showing how its performance varies as the dimensionality of the subspace is varied. The best performance for the Google dataset is with the PCA subspace dimension $d = 60$, whereas for the MSR dataset, the best performance is achieved with $d=100$. In all these cases, this experiment shows that significant reduction in the original embedding dimension can be achieved without loss of performance (in fact, with significant gains in performance).

The key difference between our approach and that proposed earlier \cite{mikolov:acl2013,goldberg-levy:2014} is the use of a relationship-specific distance metric, which is automatically learned from the given dataset, vs. using a universal relationship independent rule. Clearly, if generic rules performed extremely well across all categories, there would be no need for a relationship-specific method. Our approach is specifically designed to address the weaknesses in the "one size fits all" philosophy underlying the earlier approaches. 

\section{Future Work} 
\label{future} 


\paragraph{Relational knowledge base completion:} As discussed above, the methods tested are  related to ongoing work on relational knowledge base completion, such as {\tt TransE} \cite{bordes:aaai2011}, {\tt TransH} \cite{transh:aaai2014}, and tensor neural net methods \cite{soucher:nips2013}. The mathematical framework underlying GFK can be readily extended to relational knowledge base completion in a number of ways. First, many of these methods, like {\tt TransE} and {\tt TransH} involve finding embeddings of entities and relations that are of unit norm. For example, if a relation is modeled abstractly by a triple $(h, l, t)$, where $h$ is the head of relation $l$ and $t$ is its tail, then these embedding methods find a vector space representation for each head $h$ and tail $t$ (denoted by $\omega_h$ and $\omega_t$) such that $\|\omega_h \|_2 = \|\omega_t \|_2 = 1$.  The space of unit norm vectors defines a Grassmannian manifold, and special types of gradient methods can be developed that use the Riemannian gradient instead of the Euclidean gradient to find the suitable embedding on the Grassmannian. 

\paragraph{ Choice of Kernel:} We selected one specific kernel based on geodesic flows in this paper, but in actuality, a large number of choices for Grassmannian kernels are available for study \cite{hamm-lee:icml2008}.  These include {\em Binet-Cauchy} metric, projection metric, maximum and minimum correlation metrics, and related kernels. We are currently exploring several of these alternative choices of Grassmannian kernels for analyzing word embeddings.

\paragraph{Compact Kernel Representations:}  To address the issue of scaling our approach to large datasets, we could exploit the rich theory of representations of Lie groups, to exploit more sophisticated methods for compactly representing and efficiently computing with kernels on Lie groups.

%

\newpage 

\bibliographystyle{unsrt}

\bibliography{nips2015}

\end{document}